\documentclass[11pt,a4paper]{article}
\usepackage[T1]{fontenc}
\usepackage[utf8]{inputenc}
\usepackage{authblk}

\usepackage{acl2014}
\usepackage{times}
\usepackage{url}
\usepackage{latexsym}
\usepackage{enumitem}
\usepackage{color}
\usepackage[usenames,dvipsnames,svgnames,table]{xcolor}

\usepackage[sort,compress,square,comma,numbers]{natbib}
\usepackage[english]{babel}
\usepackage[utf8]{inputenc}
\usepackage{amsmath}
\usepackage{amsfonts}
\usepackage{graphicx}
\usepackage[T1]{fontenc}

\title{The Ubuntu Dialogue Corpus: A Large Dataset for Research in Unstructured Multi-Turn Dialogue Systems}
\author[*]{Ryan Lowe\thanks{ The first two authors contributed equally.}}
\author[*]{Nissan Pow}
\author[$\dag$]{Iulian V. Serban}
\author[*]{Joelle Pineau}
\affil[*]{School of Computer Science, McGill University, Montreal, Canada}
\affil[$\dag$]{Department of Computer Science and Operations Research, Universi{\'e} de Montr{\'e}al, Montreal, Canada}

\begin{document}

\maketitle

\begin{abstract}
This paper introduces the Ubuntu Dialogue Corpus, a dataset containing almost 1 million multi-turn dialogues, with a total of over 7 million utterances and 100 million words. This provides a unique resource for research into building dialogue managers based on neural language models that can make use of large amounts of unlabeled data.  The dataset has both the multi-turn property of conversations in the Dialog State Tracking Challenge datasets, and the unstructured nature of interactions from microblog services such as Twitter. We also describe two neural learning architectures suitable for analyzing this dataset, and provide benchmark performance on the task of selecting the best next response.
\end{abstract}

\section{Introduction}

The ability for a computer to converse in a natural and coherent manner with a human has long been held as one of the primary objectives of artificial intelligence (AI). In this paper we consider the problem of building dialogue agents that have the ability to interact in one-on-one multi-turn conversations on a diverse set of topics. We primarily target \emph{unstructured} dialogues, where there is no \emph{a priori} logical representation for the information exchanged during the conversation.  This is in contrast to recent systems which focus on structured dialogue tasks, using a slot-filling representation~\cite{williams2013dialog,henderson2014second,singh02}.

We observe that in several subfields of AI---computer vision, speech recognition, machine translation---fundamental break-throughs were achieved in recent years using machine learning methods, more specifically with neural architectures~\cite{bengio2013representation}; however, it is worth noting that many of the most successful approaches, in particular convolutional and recurrent neural networks, were known for many years prior. It is therefore reasonable to attribute this progress to three major factors:  1) the public distribution of very large rich datasets~\cite{deng2009imagenet}, 2) the availability of substantial computing power, and 3) the development of new training methods for neural architectures, in particular leveraging unlabeled data.  Similar progress has not yet been observed in the development of dialogue systems. We hypothesize that this is due to the lack of sufficiently large datasets, and aim to overcome this barrier by providing a new large corpus for research in multi-turn conversation.

The new Ubuntu Dialogue Corpus consists of almost one million two-person conversations extracted from the Ubuntu chat logs\footnote{These logs are available from 2004 to 2015 at \url{http://irclogs.ubuntu.com/}}, used to receive technical support for various Ubuntu-related problems. The conversations have an average of 8 turns each, with a minimum of 3 turns. All conversations are carried out in text form (not audio).  The dataset is orders of magnitude larger than structured corpuses such as those of the Dialogue State Tracking Challenge~\cite{williams2013dialog}.  It is on the same scale as recent datasets for solving problems such as question answering and analysis of microblog services, such as Twitter~\cite{yu2014deep,sordoni2015, shang2015neural,ritter2011data}, but each conversation in our dataset includes several more turns, as well as longer utterances.  Furthermore, because it targets a specific domain, namely technical support, it can be used as a case study for the development of AI agents in targeted applications, in contrast to chatbox agents that often lack a well-defined goal~\cite{shawar2007chatbots}.

In addition to the corpus, we present learning architectures suitable for analyzing this dataset, ranging from the simple frequency-inverse document frequency (TF-IDF) approach, to more sophisticated neural models including a Recurrent Neural Network (RNN) and a Long Short-Term Memory (LSTM) architecture. We provide benchmark performance of these algorithms, trained with our new corpus, on the task of selecting the best next response, which can be achieved without requiring any human labeling.  The dataset is ready for public release\footnote{Note that a new version of the dataset is now available: \url{https://github.com/rkadlec/ubuntu-ranking-dataset-creator}. This version makes some adjustments and fixes some bugs from the first version.}. 
The code developed for the empirical results is also available\footnote{\url{http://github.com/npow/ubottu}}.

\section{Related Work}

We briefly review existing dialogue datasets, and some of the more recent learning architectures used for both structured and unstructured dialogues. This is by no means an exhaustive list (due to space constraints), but surveys resources most related to our contribution. A list of datasets discussed is provided in Table~\ref{table:datasets}. 

\subsection{Dialogue Datasets}

The Switchboard dataset \cite{godfrey1992switchboard}, and the Dialogue State Tracking Challenge (DSTC) datasets~\cite{williams2013dialog} have been used to train and validate dialogue management systems for interactive information retrieval. The problem is typically formalized as a slot filling task, where agents attempt to predict the goal of a user during the conversation.   These datasets have been significant resources for structured dialogues, and have allowed major progress in this field, though they are quite small compared to datasets currently used for training neural architectures.

Recently, a few datasets have been used containing unstructured dialogues extracted from Twitter\footnote{\url{https://twitter.com/}}.
Ritter et al. \cite{ritter2010unsupervised} collected  1.3 million conversations; this was extended in \cite{sordoni2015} to take advantage of longer contexts by using A-B-A triples. Shang et al. \cite{shang2015neural} used data from a similar Chinese website called Weibo\footnote{\url{http://www.weibo.com/}}.  However to our knowledge, these datasets have not been made public, and furthermore, the post-reply format of such microblogging services is perhaps not as representative of natural dialogue between humans as the continuous stream of messages in a chat room. In fact, Ritter et al. estimate that only 37\% of posts on Twitter are `conversational in nature', and 69\% of their collected data contained exchanges of only length 2 \cite{ritter2010unsupervised}. We hypothesize that chat-room style messaging is more closely correlated to human-to-human dialogue than micro-blogging websites, or forum-based sites such as Reddit.

Part of the Ubuntu chat logs have previously been aggregated into a dataset, called the Ubuntu Chat Corpus \cite{uthus2013ubuntu}. However that resource preserves the multi-participant structure and thus is less amenable to the investigation of more traditional two-party conversations.

Also weakly related to our contribution is the problem of question-answer systems. Several datasets of question-answer pairs are available~\cite{boyd-graber12}, however these interactions are much shorter than what we seek to study.

\begin{table*}[t]
\centering
\scriptsize
\begin{tabular}{|l ||c|c|c|c|c|l|} \hline
Dataset & Type & Task & \# Dialogues & \# Utterances &  \# Words &  Description \\ \hline \hline

Switchboard \cite{godfrey1992switchboard}& Human-human& Various & 2,400 & ---& 3,000,000 & Telephone conversations \\ &spoken  & & & & & on pre-specified topics \\ \hline

DSTC1 \cite{williams2013dialog}& Human-computer&State& 15,000& 210,000& & Bus ride information \\ 
&spoken & tracking & & & & system \\ \hline

DSTC2 \cite{henderson2014second}& Human-computer& State  & 3,000 & 24,000 &  --- & Restaurant booking \\
& spoken & tracking & & & & system \\ \hline 

DSTC3 \cite{henderson2014dialog}& Human-computer & State  & 2,265 & 15,000& --- & Tourist information \\
& spoken & tracking & & & & system \\ \hline 

DSTC4\cite{dstc4} & Human-human & State  & 35 & --- & --- & 21 hours of tourist info  \\ 
& spoken & tracking & & & &exchange over Skype \\ \hline

\textcolor{gray}{Twitter} & \textcolor{gray}{Human-human} & \textcolor{gray}{Next utterance} & \textcolor{gray}{1,300,000} & \textcolor{gray}{3,000,000} &  --- &  \textcolor{gray}{Post/ replies extracted}  \\ 
\textcolor{gray}{Corpus} \cite{ritter2010unsupervised} & \textcolor{gray}{micro-blog} & \textcolor{gray}{generation}& & & & \textcolor{gray}{from Twitter} \\ \hline

\textcolor{gray}{Twitter Triple}& \textcolor{gray}{Human-human} & \textcolor{gray}{Next utterance} &\textcolor{gray}{ 29,000,000} & \textcolor{gray}{87,000,000} &  --- &  \textcolor{gray}{A-B-A triples from}  \\ 
\textcolor{gray}{Corpus} \cite{sordoni2015} & \textcolor{gray}{micro-blog} & \textcolor{gray}{generation}& & & & \textcolor{gray}{Twitter replies} \\ \hline

\textcolor{gray}{Sina Weibo} \cite{shang2015neural} & \textcolor{gray}{Human-human} & \textcolor{gray}{Next utterance} & \textcolor{gray}{4,435,959} & \textcolor{gray}{8,871,918} & --- & \textcolor{gray}{Post/ reply pairs extracted} \\ 
& \textcolor{gray}{micro-blog} & \textcolor{gray}{generation}& & & & \textcolor{gray}{from Weibo} \\ \hline 




\textbf{Ubuntu Dialogue} & \textbf{Human-human} & \textbf{Next utterance} & \textbf{930,000} & \textbf{7,100,000} & \textbf{100,000,000} &\textbf{Extracted from Ubuntu} \\ 
\textbf{Corpus}& \textbf{chat} & \textbf{classification} & & & & \textbf{Chat Logs} \\ \hline
\end{tabular}
\caption{A selection of structured and unstructured large-scale datasets applicable to dialogue systems. Faded datasets are not publicly available. The last entry is our contribution.} 
\label{table:datasets}
\end{table*}

\subsection{Learning Architectures}

Most dialogue research has historically focused on structured slot-filling tasks~\cite{schatzmann2005quantitative}.  Various approaches were proposed, yet few attempts leverage more recent developments in neural learning architectures.  A notable exception is the work of Henderson et al.~\cite{henderson2014word}, which proposes an RNN structure, initialized with a denoising autoencoder, to tackle the DSTC 3 domain.

Work on unstructured dialogues, recently pioneered by Ritter et al.~\cite{ritter2011data}, proposed a response generation model for Twitter data based on ideas from Statistical Machine Translation. This is shown to give superior performance to previous information retrieval (e.g.\@ nearest neighbour) approaches \cite{jafarpour2010filter}. This idea was further developed by Sordoni et al.~\cite{sordoni2015} to exploit information from a longer context, using a structure similar to the Recurrent Neural Network Encoder-Decoder model~\cite{cho2014learning}. 
This achieves rather poor performance on A-B-A Twitter triples when measured by the BLEU score (a standard for machine translation), yet performs comparatively better than the model of Ritter et al.~\cite{ritter2011data}. Their results are also verified with a human-subject study. A similar encoder-decoder framework is presented in \cite{shang2015neural}. This model uses one RNN to transform the input to some vector representation, and another RNN to `decode' this representation to a response by generating one word at a time. This model is  also evaluated in a human-subject study, although much smaller in size than in~\cite{sordoni2015}. Overall, these models highlight the potential of neural learning architectures for interactive systems, yet so far they have been limited to very short conversations.

\section{The Ubuntu Dialogue Corpus}

We seek a large dataset for research in dialogue systems with the following properties:

\begin{itemize}[noitemsep,nolistsep]
\item Two-way (or \textit{dyadic}) conversation, as opposed to multi-participant chat, preferably human-human.
\item Large number of conversations; $10^5-10^6$ is typical of datasets used for neural-network learning in other areas of AI.
\item Many conversations with several turns (more than 3).
\item Task-specific domain, as opposed to chatbot systems.
\end{itemize}
All of these requirements are satisfied by the Ubuntu Dialogue Corpus presented in this paper.

\subsection{Ubuntu Chat Logs}

The Ubuntu Chat Logs refer to a collection of logs from Ubuntu-related chat rooms on the Freenode Internet Relay Chat (IRC) network. This protocol allows for real-time chat between a large number of participants. Each chat room, or channel, has a particular topic, and  every channel participant can see all the messages posted in a given channel. Many of these channels are used for obtaining technical support with various Ubuntu issues.

As the contents of each channel are moderated, most interactions follow a similar pattern. A new user joins the channel, and asks a general question about a problem they are having with Ubuntu. Then, another more experienced user replies with a potential solution, after first addressing the 'username' of the first user. This is called a name mention \cite{uthus2013extending}, and is done to avoid confusion in the channel --- at any given time during the day, there can be between 1 and 20 simultaneous conversations happening in some channels. In the most popular channels, there is almost never a  time when only one conversation is occurring; this renders it particularly problematic to extract dyadic dialogues. A conversation between a pair of users generally stops when the problem has been solved, though some users occasionally continue to discuss a topic not related to Ubuntu.

Despite the nature of the chat room being a constant stream of messages from multiple users, it is through the fairly rigid structure in the messages that we can extract the dialogues between users. Figure \ref{fig:sample-chat} shows an example chat room conversation from the \#ubuntu channel as well as the extracted dialogues, which illustrates how users usually state the username of the intended message recipient before writing their reply (we refer to all replies and initial questions as `utterances'). For example, it is clear that users `Taru' and `kuja' are engaged in a dialogue, as are users `Old' and `bur[n]er', while user `\_pm' is asking an initial question, and `LiveCD' is perhaps elaborating on a previous comment.

\subsection{Dataset Creation}

In order to create the Ubuntu Dialogue Corpus, first a method had to be devised to extract dyadic dialogues from the chat room multi-party conversations. The first step was to separate every message into 4-tuples of (time, sender, recipient, utterance). Given these 4-tuples, it is straightforward to group all tuples where there is a matching sender and recipient. Although it is easy to separate the time and the sender from the rest, finding the intended recipient of the message is not always trivial.

\subsubsection{Recipient Identification}

While in most cases the recipient is the first word of the utterance, it is sometimes located at the end, or not at all in the case of initial questions. Furthermore, some users choose names corresponding to common English words, such as `the' or `stop', which could lead to many false positives. In order to solve this issue, we create a dictionary of usernames from the current and previous days, and compare the first word of each utterance to its entries. If a match is found, and the word does not correspond to a very common English word\footnote{We use the GNU Aspell spell checking dictionary.}, it is assumed that this user was the intended recipient of the message. If no matches are found, it is assumed that the message was an initial question, and the recipient value is left empty.

\subsubsection{Utterance Creation}

The dialogue extraction algorithm works backwards from the first response to find the initial question that was replied to, within a time frame of 3 minutes. A first response is identified by the presence of a recipient name (someone from the recent conversation history).  The initial question is identified to be the most recent utterance by the recipient identified in the first response.

All utterances that do not qualify as a first response or an initial question are discarded; initial questions that do not generate any response are also discarded. We additionally discard conversations longer than five utterances where one user says more than 80\% of the utterances, as these are typically not representative of real chat dialogues. Finally, we consider only extracted dialogues that consist of 3 turns or more to encourage the modeling of longer-term dependencies. 

To alleviate the problem of `holes' in the dialogue, where one user does not address the other explicitly, as in Figure \ref{fig:convo2}, we check whether each user talks to someone else for the duration of their conversation. If not, all non-addressed utterances are added to the dialogue.  An example conversation along with the extracted dialogues is shown in Figure \ref{fig:convo2}.  Note that we also concatenate all consecutive utterances from a given user. 

We do not apply any further pre-processing (e.g. tokenization, stemming) to the data as released in the Ubuntu Dialogue Corpus.  However the use of pre-processing is standard for most NLP systems, and was also used in our analysis (see Section~\ref{sec:learning}.)

\subsubsection{Special Cases and Limitations}

It is often the case that a user will post an initial question, and multiple people will respond to it with different answers. In this instance, each conversation between the first user and the user who replied is treated as a separate dialogue. This has the unfortunate side-effect of having the initial question appear multiple times in several dialogues. However the number of such cases is sufficiently small compared to the size of the dataset. 

Another issue to note is that the utterance posting time is not considered for segmenting conversations between two users. Even if two users have a conversation that spans multiple hours, or even days, this is treated as a single dialogue. However, such dialogues are rare.  We include the posting time in the corpus so that other researchers may filter as desired.

\subsection{Dataset Statistics}

\begin{figure}
\centering
\includegraphics[width=0.45\textwidth]{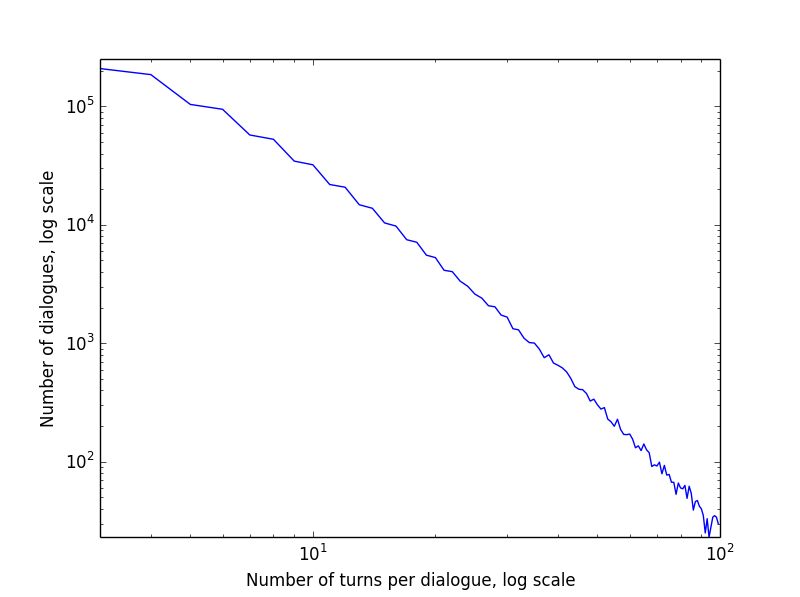}
\caption{\label{fig:turns} Plot of number of conversations with a given number of turns. Both axes use a log scale.}
\end{figure}

\begin{table}
\small
\centering
\begin{tabular}{|c |c |} \hline
\# dialogues (human-human) & 930,000  \\ \hline 
 \# utterances (in total) & 7,100,000 \\ \hline
 \# words (in total) & 100,000,000 \\ \hline
 Min. \# turns per dialogue & 3 \\ \hline 
 Avg. \# turns per dialogue & 7.71 \\ \hline
 Avg. \# words per utterance & 10.34 \\ \hline
 Median conversation length (min) & 6 \\ \hline
\end{tabular}
\caption{\label{table:dataset_stats}Properties of Ubuntu Dialogue Corpus. }
\end{table}

Table \ref{table:dataset_stats} summarizes properties of the Ubuntu Dialogue Corpus.   One of the most important features of the Ubuntu chat logs is its size. This is crucial for research into building dialogue managers based on neural architectures. Another important characteristic is the number of turns in these dialogues. The distribution of the number of turns is shown in Figure \ref{fig:turns}. It can be seen that the number of dialogues and turns per dialogue follow an approximate power law relationship.

\subsection{Test Set Generation}
\label{sec:testset}

We set aside 2\% of the Ubuntu Dialogue Corpus conversations (randomly selected) to form a test set that can be used for evaluation of response selection algorithms. Compared to the rest of the corpus, this test set has been further processed to extract a pair of \emph{(context, response, flag)} triples from each dialogue. 
The \emph{flag} is a Boolean variable indicating whether or not the response was the actual next utterance after the given context. 
The \emph{response} is a target (output) utterance which we aim to correctly identify.
The \emph{context} consists of the sequence of utterances appearing in dialogue prior to the response.
We create a pair of triples, where one triple contains the correct response (i.e. the actual next utterance in the dialogue), and the other triple contains a false response, sampled randomly from elsewhere within the test set.  The flag is set to 1 in the first case and to 0 in the second case.  An example pair is shown in Table \ref{table:triple1}.  To make the task harder, we can move from pairs of responses (one correct, one incorrect) to a larger set of wrong responses (all with flag=0).  In our experiments below, we consider both the case of 1 wrong response and 10 wrong responses.

\begin{table}[!ht]
\scriptsize
\begin{tabular}{|l |l |c|} \hline
Context & Response & Flag \\ \hline
well, can I move the drives? & I guess I could just  & 1 \\ \_\_EOS\_\_ ah not like that & get an enclosure and  & \\ 
& copy via USB & \\ \hline
well, can I move the drives? & you can use "ps ax"  & 0 \\ \_\_EOS\_\_ ah not like that & and "kill (PID \#)"  & \\ \hline
\end{tabular}
\caption{\label{table:triple1}Test set example with (context, reply, flag) format. The '\_\_EOS\_\_' tag is used to denote the end of an utterance within the context.} 
\end{table}

Since we want to learn to predict all parts of a conversation, as opposed to only the closing statement, we consider various portions of context for the conversations in the test set. The context size is determined stochastically using a simple formula:
$$
c = \min (t - 1, n - 1),
$$ $$
\textmd{where } \textmd{ } n = \frac{10C}{\eta} + 2, \eta \sim Unif(C/2,10C)
$$
Here, $C$ denotes the maximum desired context size, which we set to $C = 20$. The last term is the desired minimum context size, which we set to be 2. Parameter $t$ is the actual length of that dialogue (thus the constraint that $c \leq t-1$), and $n$ is a random number corresponding to the randomly sampled context length, that is selected to be inversely proportional to C.

In practice, this leads to short test dialogues having short contexts, while longer dialogues are often broken into short or medium-length segments, with the occasional long context of 10 or more turns.


\subsection{Evaluation Metric}

We consider the task of best response selection.   This can be achieved by processing the data as described in Section~\ref{sec:testset}, without requiring any human labels.     This classification task is an adaptation of the recall and precision metrics previously applied to dialogue datasets~\cite{schatzmann2005quantitative}.

A family of metrics often used in language tasks is Recall@k (denoted R@1 R@2, R@5 below).  Here the agent is asked to select the $k$ most likely responses, and it is correct if the true response is among these $k$ candidates.  Only the R@1 metric is relevant in the case of binary classification (as in the Table~\ref{table:triple1} example).

Although a language model that performs well on response classification is not a gauge of good performance on next utterance generation, we hypothesize that improvements on a model with regards to the classification task will eventually lead to improvements for the generation task.  See Section~\ref{sec:discussion} for further discussion of this point.

\section{Learning Architectures for Unstructured Dialogues}
\label{sec:learning}
To provide further evidence of the value of our dataset for research into neural architectures for dialogue managers, we provide performance benchmarks for two neural learning algorithms, as well as one naive baseline. The approaches considered are: TF-IDF, Recurrent Neural networks (RNN), and Long Short-Term Memory (LSTM). Prior to applying each method, we perform standard pre-processing of the data using the NLTK\footnote{\url{www.nltk.org/}} library and Twitter tokenizer\footnote{\url{http://www.ark.cs.cmu.edu/TweetNLP/}} to parse each utterance. We use generic tags for various word categories, such as names, locations, organizations, URLs, and system paths.
  
To train the RNN and LSTM architectures, we process the full training Ubuntu Dialogue Corpus into the same format as the test set described in Section~\ref{sec:testset}, extracting \emph{(context, response, flag)} triples from dialogues.  For the training set, we do not sample the context length, but instead consider each utterance (starting at the 3rd one) as a potential response, with the previous utterances as its context.  So a dialogue of length 10 yields 8 training examples. Since these are overlapping, they are clearly not independent, but we consider this a minor issue given the size of the dataset (we further alleviate the issue by shuffling the training examples).  Negative responses are selected at random from the rest of the training data.

\subsection{TF-IDF}

Term frequency-inverse document frequency is a statistic that intends to capture how important a given word is to some document, which in our case is the context \cite{ramos2003using}. It is a technique often used in document classification and information retrieval. The `term-frequency' term is simply a count of the number of times a word appears in a given  context, while the `inverse document frequency' term puts a penalty on how often the word appears elsewhere in the corpus. The final score is calculated as the product of these two terms, and has the form:
$$
\textmd{tfidf}(w,d,D) = f(w,d) \times \log \frac{N}{|\{d \in D : w \in d\}|},
$$
where $f(w,d)$ indicates the number of times word $w$ appeared in context $d$, $N$ is the total number of dialogues, and the denominator represents the number of dialogues in which the word $w$  appears. 

For classification, the TF-IDF vectors are first calculated for the context and each of the candidate responses. Given a set of candidate response vectors, the one with the highest cosine similarity to the context vector is selected as the output. For Recall@k, the top k responses are returned.

\subsection{RNN}

Recurrent neural networks are a variant of neural networks that allows for time-delayed directed cycles between units \cite{medsker2001recurrent}. This leads to the formation of an internal state of the network, $h_t$,  which allows it to model time-dependent data. The internal state is updated at each time step as some function of the observed variables $x_t$, and the hidden state at the previous time step $h_{t-1}$. $W_{x}$ and $W_{h}$ are matrices associated with the input and hidden state.

$$ h_{t} = f(W_{h}h_{t-1} + W_{x}x_t).$$

\begin{figure}
\centering
\includegraphics[width=0.85\linewidth]{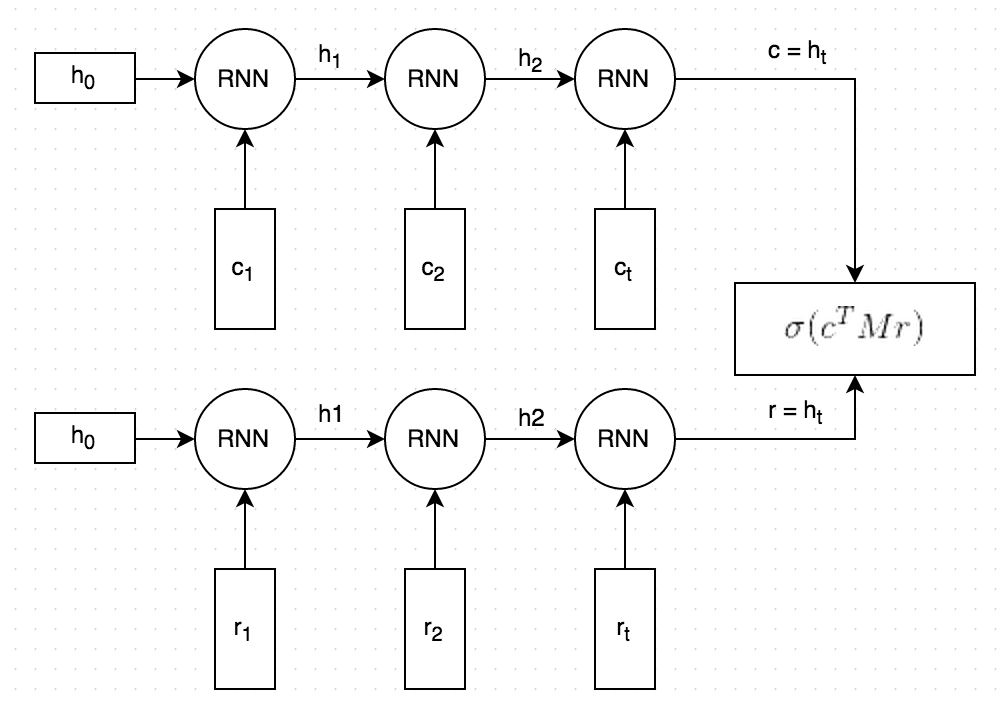}
\caption{\label{rnn1}Diagram of our model. The RNNs have tied weights. $c, r$ are the last hidden states from the RNNs. $c_{i}, r_{i}$ are word vectors for the context and response, $i < t$. We consider contexts up to a maximum of $t = 160$. }
\end{figure}

A diagram of an RNN can be seen in Figure~\ref{rnn1}. RNNs have been the primary building block of many current neural language models \cite{ritter2011data, sordoni2015}, which use RNNs for an encoder and decoder. The first RNN is used to encode the given context, and the second RNN generates a response by using beam-search, where its initial hidden state is biased using the final hidden state from the first RNN.  In our work, we are concerned with classification of responses, instead of generation. We build upon the approach in \cite{bordes2014open}, which has also been recently applied to the problem of question answering~\cite{yu2014deep}.

We utilize a siamese network consisting of two RNNs with tied weights to produce the embeddings for the context and response. Given some input context and response, we compute their embeddings --- $c, r \in \mathbb{R}^d$, respectively --- by feeding the word embeddings one at a time into its respective RNN. Word embeddings are initialized using the pre-trained vectors (Common Crawl, 840B tokens from \cite{Pennington2014}), and fine-tuned during training. The hidden state of the RNN is updated at each step, and the final hidden state represents a \emph{summary} of the input utterance. Using the final hidden states from both RNNs, we then calculate the probability that this is a valid pair:
$$
p(\textmd{flag}=1|c,r,M) = \sigma(c^T M r + b),
$$
where the bias $b$ and the matrix $M \in \mathbb{R}^{d\times d}$ are learned model parameters. This can be thought of as a generative approach; given some input response, we generate a context with the product $c' = Mr$, and measure the similarity to the actual context using the dot product. This is converted to a probability with the sigmoid function. The model is trained by minimizing the cross entropy of all labeled (context, response) pairs \cite{yu2014deep}:
$$
\mathcal{L} = - \sum_n \log p(\textmd{flag}_n|c_n,r_n,M) + \frac{\lambda}{2}||\theta=||_F^2
$$
where $||\theta||_F^2$ is the Frobenius norm of $\theta = \{M,b\}$. In our experiments, we use $\lambda=0$ for computational simplicity.

For training, we used a 1:1 ratio between true responses (flag = 1), and negative responses (flag=0) drawn randomly from elsewhere in the training set. The RNN architecture is set to 1 hidden layer with 50 neurons. The $W_{h}$ matrix is initialized using orthogonal weights \cite{saxe2013exact}, while $W_{x}$ is initialized using a uniform distribution with values between -0.01 and 0.01. We use Adam as our optimizer \cite{DBLP:journals/corr/KingmaB14}, with gradients clipped to 10. We found that weight initialization as well as the choice of optimizer were critical for training the RNNs.


\subsection{LSTM}

In addition to the RNN model, we consider the same architecture but changed the hidden units to long-short term memory (LSTM) units \cite{hochreiter1997long}. LSTMs were introduced in order to model longer-term dependencies. This is accomplished using a series of gates that determine whether a new input should be remembered, forgotten (and the old value retained), or used as output. The error signal can now be fed back indefinitely into the gates of the LSTM unit. This helps overcome the vanishing and exploding gradient problems in standard RNNs, where the error gradients would otherwise decrease or increase at an exponential rate. In training, we used 1 hidden layer with 200 neurons.  The hyper-parameter configuration (including number of neurons) was optimized independently for RNNs and LSTMs using a validation set extracted from the training data.

\section{Empirical Results}

The results for the TF-IDF, RNN, and LSTM models are shown in Table \ref{results}. The models were evaluated using both 1 (1 in 2) and 9 (1 in 10) false examples.  Of course, the Recall@2 and Recall@5 are not relevant in the binary classification case\footnote{Note that these results are on the original dataset. Results on the new dataset should not be compared to the old dataset; baselines on the new dataset will be released shortly.}.

\begin{table}[!ht]
\centering
\begin{tabular}{|c |c |c|c|} \hline
Method & TF-IDF & RNN & LSTM \\ \hline \hline
1 in 2 R@1 & 65.9\% & 76.8\% & \textbf{87.8\%} \\ \hline 
1 in 10 R@1 & 41.0\% & 40.3\% & \textbf{60.4\%} \\ \hline
1 in 10 R@2 & 54.5\% & 54.7\% & \textbf{74.5\%} \\ \hline 
1 in 10 R@5 & 70.8\% & 81.9\% & \textbf{92.6\%} \\ \hline
\end{tabular}
\caption{\label{results}Results for the three algorithms using various recall measures for binary (1 in 2) and 1 in 10 (1 in 10) next utterance classification \%.}
\end{table}

We observe that the LSTM outperforms both the RNN and TF-IDF on all evaluation metrics. It is interesting to note that TF-IDF actually outperforms the RNN on the Recall@1 case for the 1 in 10 classification. This is most likely due to the limited ability of the RNN to take into account long contexts, which can be overcome by using the LSTM. An example output of the LSTM where the response is correctly classified is shown in Table \ref{example}.

\begin{figure}
\centering
\includegraphics[width=0.45\textwidth]{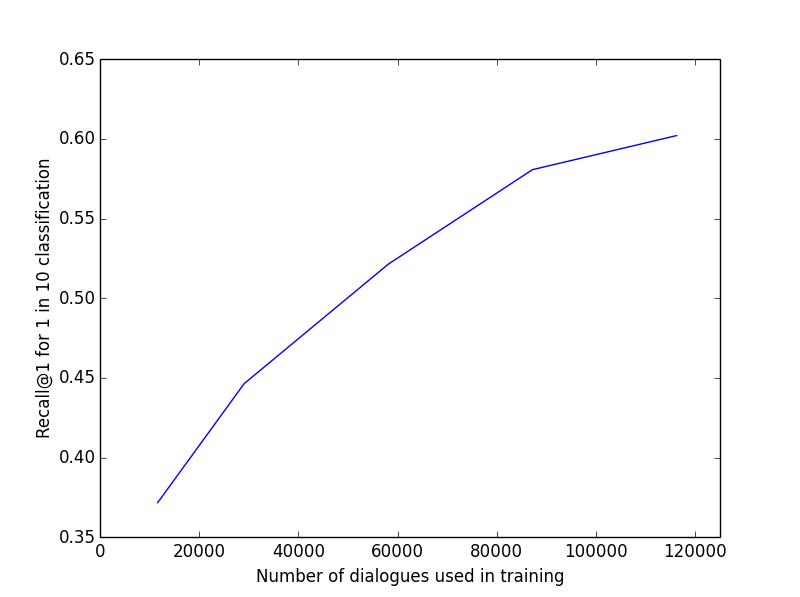}
\caption{\label{datasize} The LSTM (with 200 hidden units), showing Recall@1 for the 1 in 10 classification, with increasing dataset sizes.}
\end{figure}

We also show, in Figure \ref{datasize}, the increase in performance of the LSTM as the amount of data used for training increases. This confirms the importance of having a large training set.

\begin{table}[!ht]
\small
\begin{tabular}{|l|} \hline
Context \\ \hline
""any apache hax around ? i just deleted all of \\
\_\_path\_\_ - which package provides it ?", \\ 
"reconfiguring apache do n't solve it ?" \\ \hline \hline
\end{tabular}
\begin{tabular}{|l |c|} \hline 
Ranked Responses & Flag \\ \hline
1. "does n't seem to, no" & 1  \\ \hline
2. "you can log in but not transfer files ?"  & 0  \\ \hline 
\end{tabular}
\caption{\label{example}Example showing the ranked responses from the LSTM. Each utterance is shown after pre-processing steps.}
\end{table}

\section{Discussion}
\label{sec:discussion}

This paper presents the Ubuntu Dialogue Corpus, a large dataset for research in unstructured multi-turn dialogue systems.  We describe the construction of the dataset and its properties.   The availability of a dataset of this size opens up several interesting possibilities for research into dialogue systems based on rich neural-network architectures.  We present preliminary results demonstrating use of this dataset to train an RNN and an LSTM  for the task of selecting the next best response in a conversation; we obtain significantly better results with the LSTM architecture. 
There are several interesting directions for future work.

\subsection{Conversation Disentanglement}

Our approach to conversation disentanglement consists of a small set of rules.  More sophisticated techniques have been proposed, such as training a maximum-entropy classifier to cluster utterances into separate dialogues~\cite{elsner2008you}. 
However, since we are not trying to replicate the \textit{exact} conversation between two users, but only to retrieve \textit{plausible} natural dialogues, the heuristic method presented in this paper may be sufficient. This seems supported through qualitative examination of the data, but could be the subject of more formal evaluation. 

\subsection{Altering Test Set Difficulty}

One of the interesting properties of the response selection task is the ability to alter the task difficulty in a controlled manner. We demonstrated this by moving from 1 to 9 false responses, and by varying the Recall@k parameter.  In the future, instead of choosing false responses randomly, we will consider selecting false responses that are similar to the actual response (e.g. as measured by cosine similarity). A dialogue model that performs well on this more difficult task should also manage to capture a more fine-grained semantic meaning of sentences, as compared to a model that naively picks replies with the most words in common with the context such as TF-IDF.

\subsection{State Tracking and Utterance Generation}

The work described here focuses on the task of response selection. This can be seen as an intermediate step between slot filling and utterance generation.  In slot filling, the set of candidate outputs (\emph{states}) is identified \emph{a priori} through knowledge engineering, and is typically smaller than the set of responses considered in our work.  
When the set of candidate responses is close to the size of the dataset (e.g. all utterances ever recorded), then we are quite close to the response generation case.


There are several reasons not to proceed directly to response generation.  First, it is likely that current algorithms are not yet able to generate good results for this task, and it is preferable to tackle metrics for which we can make progress.  Second, 
we do not yet have a suitable metric for evaluating performance in the response generation case.  
One option is to use the BLEU \cite{papineni2002bleu} or METEOR \cite{Lavie2009} scores from machine translation. However, using BLEU to evaluate dialogue systems has been shown to give extremely low scores \cite{sordoni2015}, due to the large space of potential sensible responses \cite{galley2015deltableu}.   Further, since the BLEU score is calculated using N-grams \cite{papineni2002bleu}, it would provide a very low score for reasonable responses that do not have any words in common with the ground-truth next utterance.

Alternatively, one could measure the difference between the generated utterance and the actual sentence by comparing their representations in some embedding (or \emph{semantic}) space.  However, different models inevitably use different embeddings, necessitating a standardized embedding for evaluation purposes. Such a standardized embeddings has yet to be created. 

Another possibility is to use human subjects to score automatically generated responses, but time and expense make this a highly impractical option.

In summary, while it is possible that current language models have outgrown the use of slot filling as a metric, we are currently unable to measure their ability in next utterance generation in a standardized, meaningful and inexpensive way.  This motivates our choice of response selection as a useful metric for the time being.

\section*{Acknowledgments}

The authors gratefully acknowledge financial support for this work by the Samsung Advanced Institute of Technology (SAIT) and the Natural Sciences and Engineering Research Council of Canada (NSERC).  We would like to thank Laurent Charlin for his input into this paper, as well as Gabriel Forgues and Eric Crawford for interesting discussions. 


\bibliographystyle{plain}
\bibliography{bibliography.bib}

\nocite{zeiler2012adadelta}
\nocite{wang2013dataset}

\section*{Appendix A: Dialogue excerpts}

\begin{figure}[h!]
\centering
\scriptsize
\begin{tabular}{|c |c |l|} \hline
Time & User & Utterance \\ \hline \hline
03:44& Old& I dont run graphical ubuntu,\\&& I run ubuntu server. \\ \hline
03:45& kuja& Taru: Haha sucker.\\ \hline
03:45& Taru &Kuja: ?\\ \hline
03:45& bur[n]er& Old: you can use "ps ax" \\&&and "kill (PID\#)"\\ \hline
03:45& kuja& Taru: Anyways, you made \\&& the changes right?\\ \hline
03:45& Taru& Kuja: Yes.\\ \hline
03:45& LiveCD& or killall speedlink\\ \hline
03:45& kuja& Taru: Then from the terminal\\&& type:  sudo apt-get update\\ \hline
03:46& \_pm& if i install the beta version, \\&& how can i update it when \\&& the final version comes out? \\ \hline
03:46& Taru& Kuja: I did.\\ \hline 
\end{tabular}
\begin{tabular}{|c |c |l|}  \hline
Sender & Recipient & Utterance \\ \hline \hline
Old& & I dont run graphical ubuntu,\\&& I run ubuntu server. \\ \hline
bur[n]er& Old& you can use "ps ax" and \\&& "kill (PID\#)"\\ \hline \hline
kuja& Taru& Haha sucker.\\ \hline
 Taru &Kuja& ?\\ \hline
kuja& Taru& Anyways, you made the \\&& changes right?\\ \hline
Taru& Kuja& Yes.\\ \hline
kuja& Taru& Then from the terminal type:\\&&   sudo apt-get update\\ \hline
Taru& Kuja& I did.\\ \hline 
\end{tabular}
\caption{\label{fig:sample-chat}Example chat room conversation from the \#ubuntu channel of the Ubuntu Chat Logs (top), with the disentangled conversations for the Ubuntu Dialogue Corpus (bottom).}
\end{figure}

\begin{figure}[h!]
\centering
\scriptsize
\begin{tabular}{|c |c |l|} \hline
Time & User & Utterance \\ \hline \hline
[12:21] & dell & well, can I move the drives? \\ \hline
[12:21] &cucho& dell: ah not like that \\ \hline 
[12:21] &RC& dell: you can't move the drives \\ \hline 
[12:21] &RC& dell: definitely not\\ \hline 
[12:21] &dell &ok\\ \hline 
[12:21] &dell& lol\\ \hline 
[12:21] &RC &this is the problem with RAID:)\\ \hline 
[12:21] &dell & RC haha yeah \\ \hline
[12:22] &dell& cucho, I guess I could \\ &&just get   an enclosure \\ && and copy via USB...\\ \hline 
[12:22] &cucho& dell: i would advise you to get\\&& the disk\\ \hline 
\end{tabular}
\begin{tabular}{|c |c |l|} \hline
Sender & Recipient & Utterance \\ \hline \hline
dell & & well, can I move the drives? \\ \hline 
cucho& dell & ah not like that \\ \hline
dell & cucho & I guess I could just get an \\ && enclosure and copy via USB \\ \hline
cucho & dell & i would advise you to get the \\&& disk \\ \hline \hline
dell & & well, can I move the drives? \\ \hline 
RC & dell & you can't move the drives. \\&& definitely not. this is \\&& the problem with RAID :) \\ \hline
dell & RC & haha yeah \\ \hline
\end{tabular}
\caption{\label{fig:convo2}Example of before (top box) and after (bottom box) the algorithm adds and concatenates utterances in dialogue extraction. Since \texttt{RC} only addresses \texttt{dell}, all of his utterances are added, however this is not done for \texttt{dell} as he addresses both \texttt{RC} and \texttt{cucho}. }
\end{figure}

\end{document}